%% file: main.tex
\documentclass[lettersize,journal]{IEEEtran}

\usepackage[top=0.75in, bottom=1.05in, left=0.75in, right=0.75in]{geometry}
\usepackage{amsmath,amssymb,amsfonts}
\usepackage{graphicx}
\usepackage{pifont}
\usepackage{tikz}

\usepackage{amsthm}
\usepackage{amsmath}
\usepackage{xcolor}
\usepackage[figuresright]{rotating}
\usepackage[colorlinks=true, hidelinks, bookmarks=false]{hyperref}

\usepackage{graphicx}
\graphicspath{{/}{fig/}}

\usepackage{array}
\usepackage{textcomp}
\usepackage{xcolor}
\usepackage{multirow}
\usepackage{booktabs}

\usepackage{enumitem}

\usepackage{mathtools}

\usepackage{float}

\usepackage{fancyhdr}

\usepackage{pgfplots}
\pgfplotsset{compat=1.7}
\usepgfplotslibrary{groupplots}

\usepackage{silence}
\usepackage{orcidlink}
\usepackage{tabularx}
\usepackage{flushend}
\usepackage{xpatch}
\usepackage{algorithmic}
\usepackage[linesnumbered,ruled,vlined,shortend]{algorithm2e}
\usepackage{subcaption}
\hypersetup{draft}  

\newlength\figureheight
\newlength\figurewidth
\SetAlCapNameFnt{\footnotesize}
\SetAlCapFnt{\footnotesize}

\definecolor{bestcolor}{RGB}{255, 69, 0} 
\definecolor{secondcolor}{RGB}{255, 165, 0} 

\title{
TriLICT: Native LiDAR Image and Camera Fusion for Robust Semantic Segmentation
}

\author{

        Haizhou Zhang\,\orcidlink{0009-0005-1321-8687},
        Ha Sier\,\orcidlink{0009-0000-3617-107X},
        Jiaqiang Zhang\,\orcidlink{0000-0002-4509-8115},
        Ali Salmasi\,\orcidlink{0009-0001-3982-9962},
        Xianjia Yu\,\orcidlink{0000-0002-9042-3730},
        Tomi Westerlund\,\orcidlink{0000-0002-1793-2694}

\thanks{Haizhou Zhang and Xianjia Yu are the corresponding authors.}
\thanks{This research is supported by the Research Council of Finland's Digital Waters (DIWA) flagship (Grant No. 359247) and the DIWA Doctoral Training Pilot project funded by the Ministry of Education and Culture (Finland).}
\thanks{All authors are with \href{https://tiers.utu.fi}{Turku Intelligent Embedded and Robotic Systems (TIERS) Lab}, University of Turku, Turku, Finland.(e-mail:\{hazhan, sier.s.ha, jiaqiang.zhang,ali.a.salmasi,  xianjia.yu, tovewe\}@utu.fi)}

}

\fancypagestyle{FirstPage}{

}

\begin{document}

\maketitle

\input{sec/00_Abstract}
\IEEEpeerreviewmaketitle

\input{sec/01_Introduction}
\input{sec/02_RelatedWorks}
\input{sec/03_Methodology}
\input{sec/04_Experiment}
\input{sec/05_Results}
\input{sec/06_Conclusion}


\bibliographystyle{unsrt}
\bibliography{bibliography}


\makeatletter
\xpatchcmd{\IEEEbiography}{plus 1fil}{plus 0pt}{}%
  {\PackageWarning{main}{IEEEbiography patch failed; bio gaps will stretch}}
\makeatother
\raggedend
\raggedbottom

\vspace{-25pt}

\begin{IEEEbiography}
[{\includegraphics[width=1in,height=1.25in,clip,keepaspectratio]{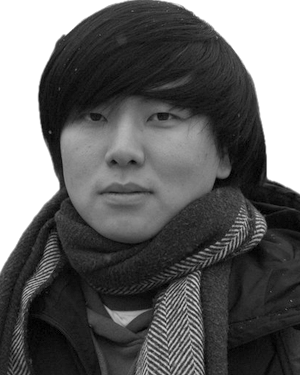}}]{Haizhou Zhang} received his B.Eng. degree in Telecommunication Engineering from Sun Yat-Sen University, China, in 2020, and the M.Sc. degree in Information and Communication Technology from the University of Turku, Finland, in 2024 (jointly with Fudan University, China). He is currently working toward the D.Sc. degree in Robotics at the University of Turku. His main research interests include multimodal sensor fusion and collaborative robot systems, with a focus on deep learning-based perception methods.

\end{IEEEbiography}

\vspace{-25pt}

\begin{IEEEbiography}
[{\includegraphics[width=1in,height=1.25in,clip,keepaspectratio]{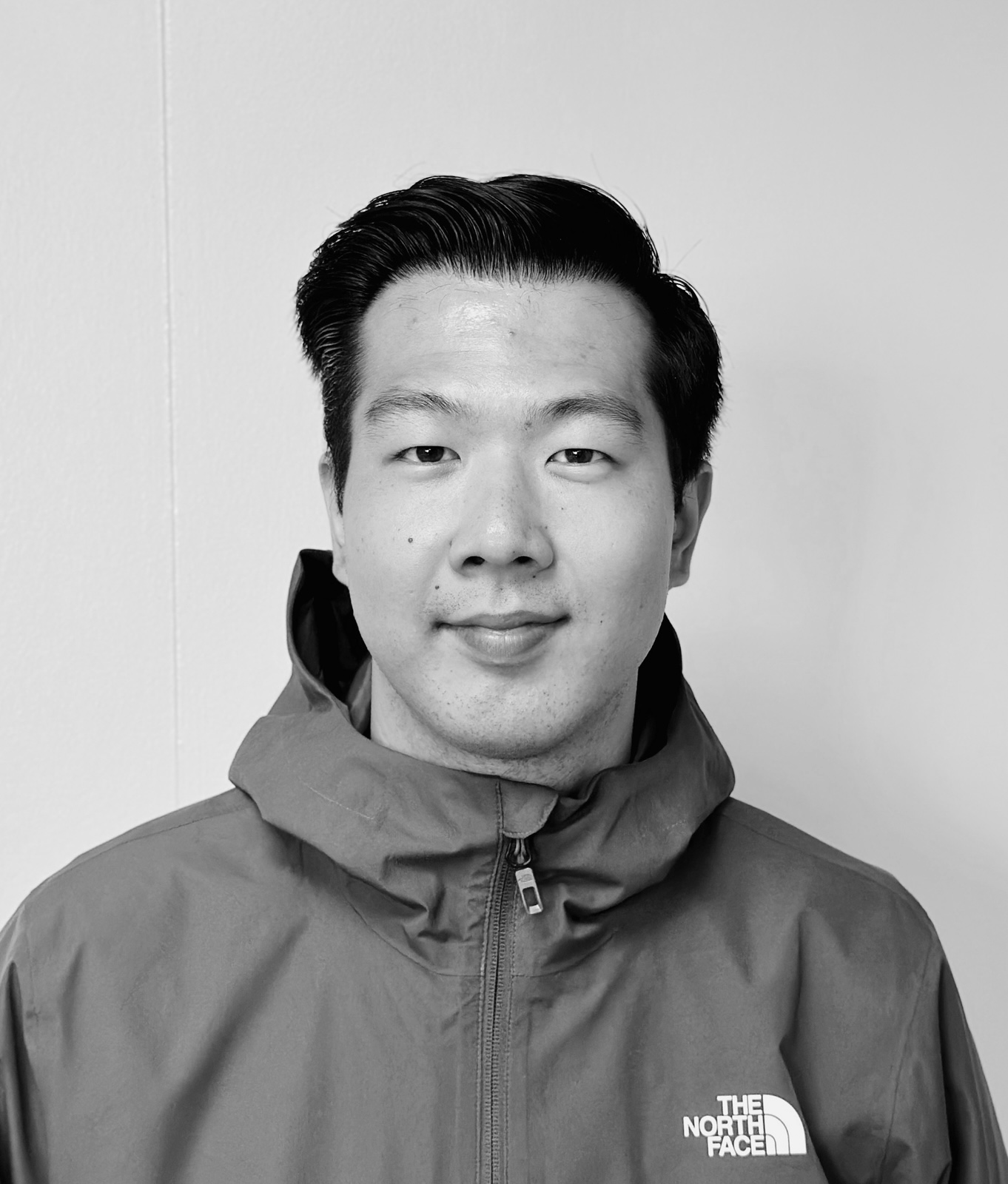}}]{Ha Sier} received the B.Eng. degree in electronic information engineering from Northeastern University, Shenyang, China, in 2019, and the M.Sc. degree in information and communication technology from the University of Turku, Turku, Finland, in 2023, jointly awarded with Fudan University, Shanghai, China.

He is currently pursuing the Ph.D. degree in robotics with the Turku Intelligent Embedded and Robotic Systems (TIERS) Lab, University of Turku. His main research interests include multi-modal sensor fusion, LiDAR-based SLAM, and robust perception for autonomous mobile robots in unstructured and GNSS-denied environments.
\end{IEEEbiography}

\vspace{-25pt}

\begin{IEEEbiography}
[{\includegraphics[width=1in,height=1.25in,clip,keepaspectratio]{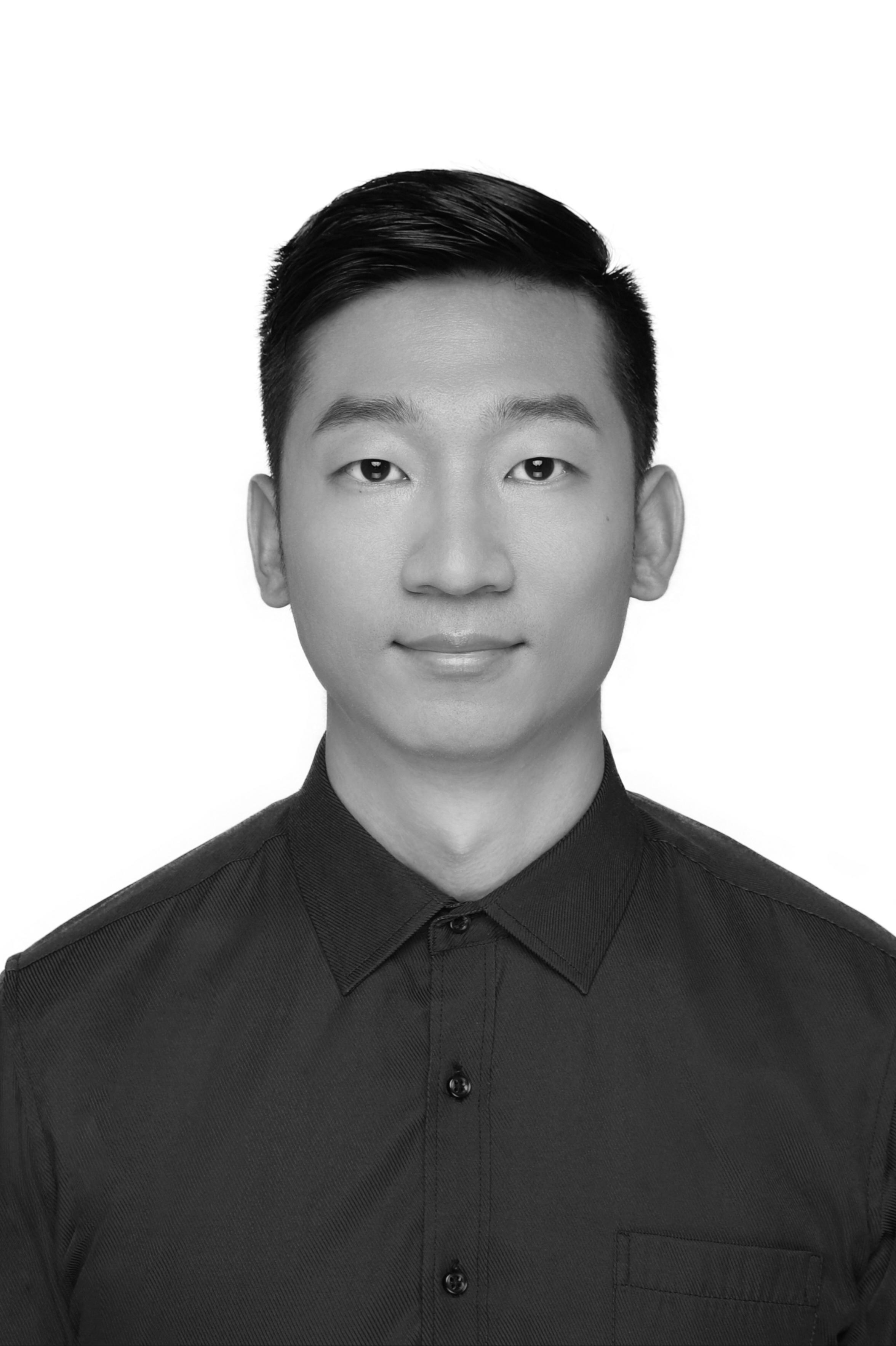}}]{Jiaqiang Zhang} received his Ph.D. in Information and Communication Technology from the University of Turku, Turku, Finland, in 2026. He holds a Master of Science in Circuits and Systems from the Shanghai Institute of Technical Physics of the Chinese Academy of Sciences, Shanghai, P.R.China, obtained in 2020. Previously, he worked as a software engineer in Shanghai, China. His research interests include robotics, multi-modal sensing and perception, sensor fusion, and remote sensing.
\end{IEEEbiography}

\vspace{-25pt}

\begin{IEEEbiography}
[{\includegraphics[width=1in,height=1.25in,clip,keepaspectratio]{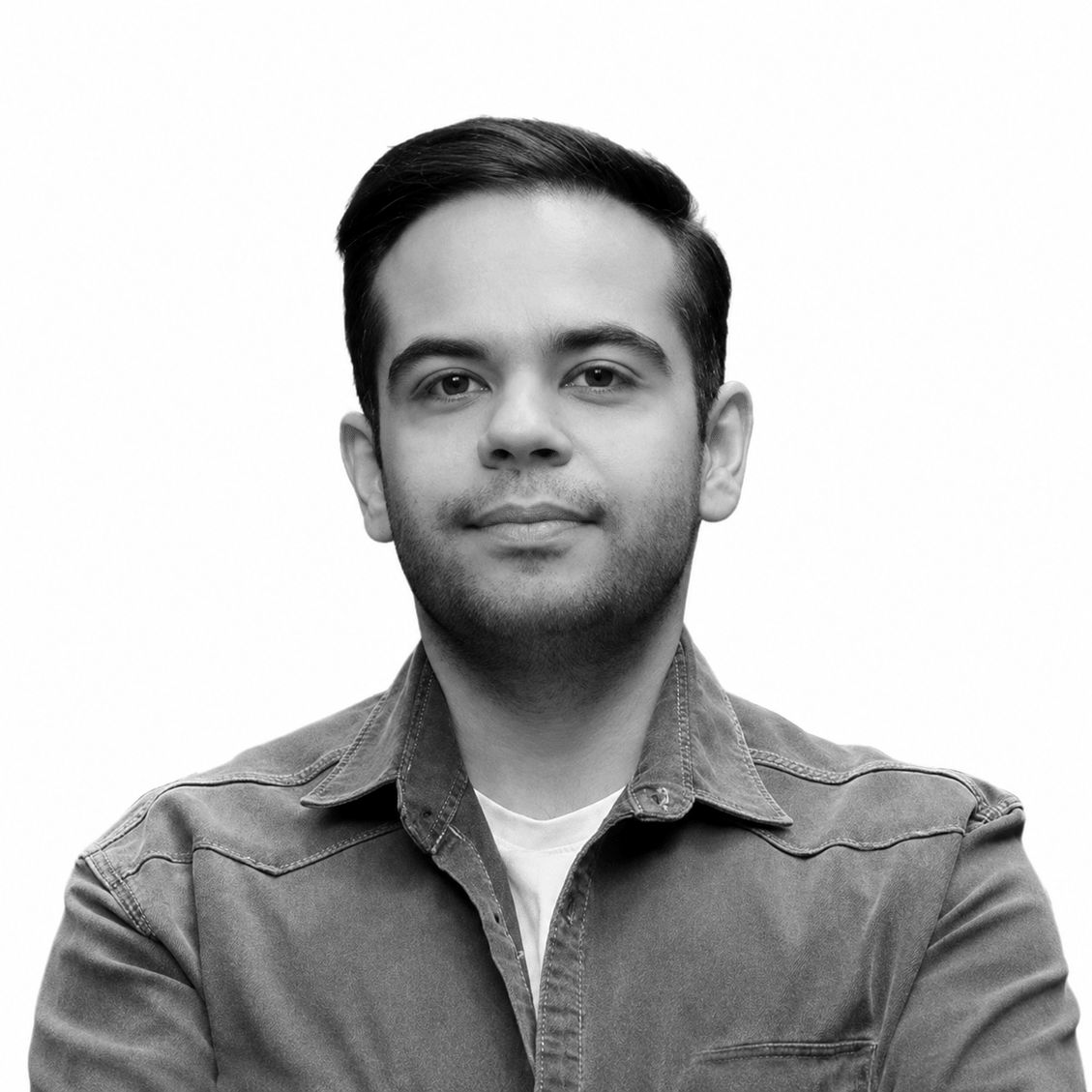}}]{Ali Salmasi} received the B.Sc. degree in electrical engineering in 2019, and the M.Sc. degree in electrical engineering from Tarbiat Modares University, Tehran, Iran, in 2023. He is currently pursuing the Ph.D. degree in robotics with the Turku Intelligent Embedded and Robotic Systems (TIERS) Lab, University of Turku, Turku, Finland. His main research interests include computer vision, multimodal sensor fusion, vision-language models, and autonomous robot navigation, with a focus on learning-based perception and semantic for understanding autonomous and collaborative robotic systems.
\end{IEEEbiography}

\vspace{-25pt}

\begin{IEEEbiography}
[{\includegraphics[width=1in,height=1.25in,clip,keepaspectratio]{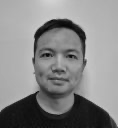}}]{Xianjia Yu} received the D.Sc. (Tech.) degree in Robotics and Autonomous Systems from the University of Turku, Turku, Finland, in 2024. He is currently a Postdoctoral Researcher with the Turku Intelligent Embedded and Robotic Systems (TIERS) Group, University of Turku, and an Automation Researcher and Engineer with Kaptas Oy, Turku, Finland. He has been engaged in robotics research and development since 2018, with experience spanning autonomous mobile robots, multimodal perception and sensor fusion, localization and mapping, and industrial robotic systems. His research interests include multimodal sensing and perception, machine learning for robotics, multi-robot systems, and real-world robotic and automation systems.
\end{IEEEbiography}

\vspace{-25pt}

\begin{IEEEbiography}
[{\includegraphics[width=1in,height=1.25in,clip,keepaspectratio]{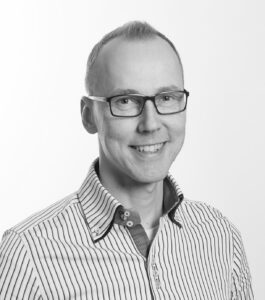}}]{Tomi Westerlund} is a professor in robotics and autonomous systems. He is the research group leader of the Turku Intelligent Embedded and Robotics Systems (TIERS) lab at the University of Turku (UTU), Finland. His main research interests are in collaborative and heterogeneous multi-robot systems and autonomous robots in urban and unstructured environments. In essence, to better understand how robots operate and can perform a variety of functions in different fields like environmental monitoring, delivery, and search and rescue missions. Westerlund has over 220 peer-reviewed publications. He is a senior member of IEEE.
\end{IEEEbiography}


\end{document}

%% file: sec/00_Abstract.tex

\begin{abstract}\label{sec:abstract}

Rapid advancements in automotive LiDAR hardware, exemplified by its custom L3 digital LiDAR chip, enable high-resolution, multi-channel native LiDAR images(e.g., range, signal, reflectivity, and near-infrared), moving beyond traditional 3D-point-to-2D-image projections. As LiDAR is an active sensing modality, these representations are largely invariant to ambient illumination, providing complementary information when camera imagery is degraded under nighttime conditions. However, how to best exploit these native LiDAR images for dense prediction and fuse them with RGB imagery, remains under-explored. Hence, we present TriLICT(Tri-LiDAR Image and Camera Transformer), a dual-stream transformer framework which employs parallel encoders to process a three-channel LiDAR image alongside a camera-based RGB input for semantic segmentation. We systematically train and evaluate independent models for different modality combinations and sensing scenarios. Experiments across daytime and nighttime conditions show that our dual-fusion model achieves substantial improvements over the conventional RGB baseline in nighttime scenarios, increasing mIoU from 52.6\% to 74.0\% with a ViT-Base/16 backbone while matching the baseline in daytime. 
To facilitate further research in multimodal perception using native multi-channel LiDAR imagery, we also contribute a publicly available annotated semantic segmentation dataset featuring synchronized RGB and multi-channel LiDAR images.

\end{abstract}


%% file: sec/01_Introduction.tex
\section{Introduction}\label{sec:introduction}
Among the primary perception modalities, RGB cameras provide dense, texture-rich imagery that carries strong semantic information, but they are sensitive to lighting conditions and perform poorly in nighttime or low-illumination environments~\cite{ahn2019analysis}. LiDARs, in contrast, deliver precise three-dimensional geometric measurements that are more robust to illumination changes~\cite{felix2025understanding}. The complementary characteristics of these modalities have therefore motivated extensive research on camera--LiDAR fusion for autonomous driving~\cite{liu2023real, caltagirone2019lidar}.

Recent advances in digital LiDAR hardware further expand the information available for perception. For example, Ouster's OS-series sensors, powered by its custom L3 digital LiDAR system-on-chip (SoC), provide high-resolution sensing designed for applications including automotive and autonomous operations~\cite{OusterREV7}.
In addition to range measurements, Ouster sensors natively provide data channels including range, signal strength, near-infrared intensity, and reflectivity. Unlike image representations constructed through a simple projection of 3D point clouds onto a 2D image plane by post processing, these structured measurements are available natively from the sensor and can be directly processed in image form.

\begin{figure}[!t]
  \centering
  \includegraphics[width=0.96\columnwidth]{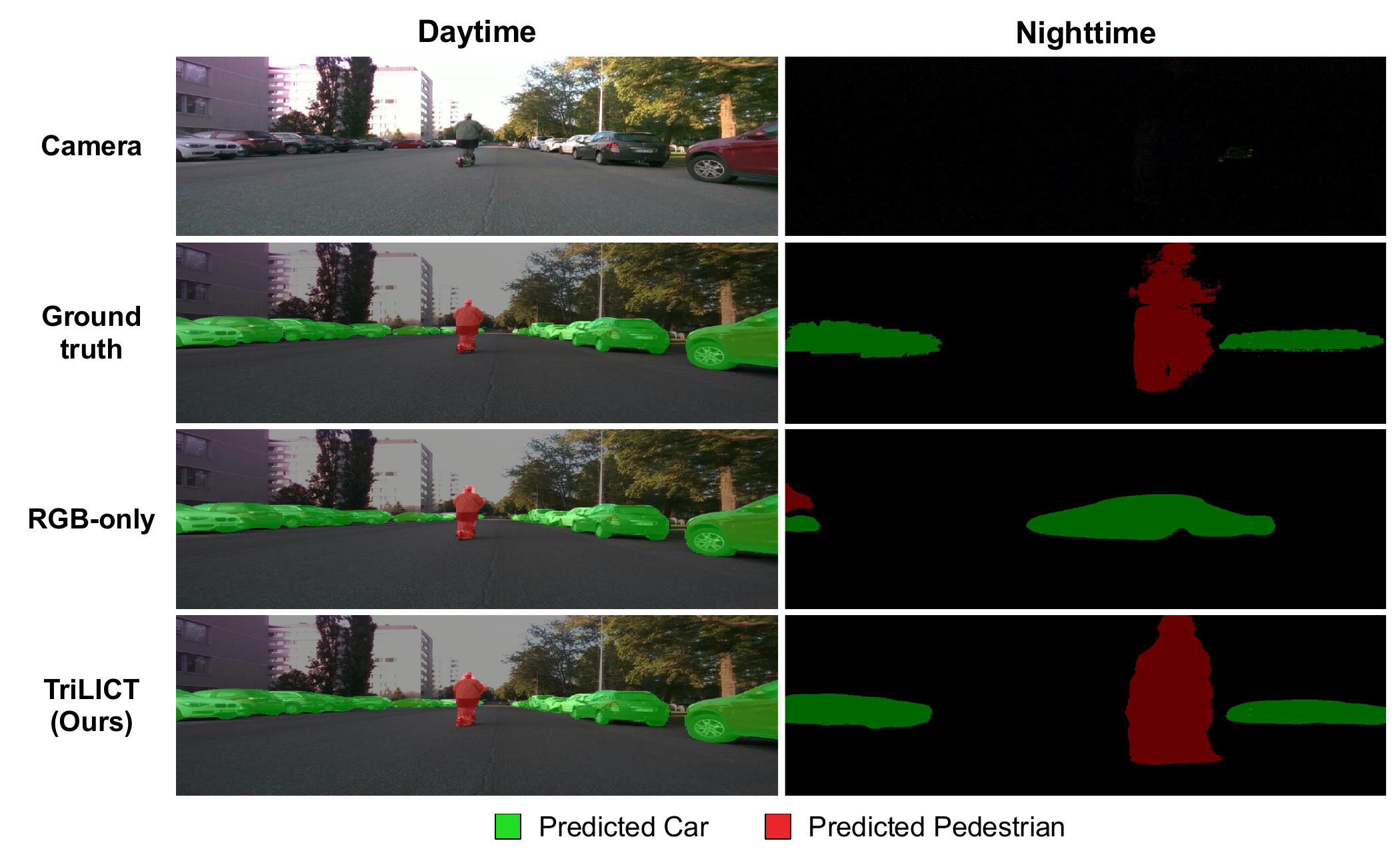}
  \caption{Qualitative comparison of RGB-only and TriLICT semantic segmentation under daytime and nighttime conditions. Despite extremely low illumination at night, TriLICT successfully segments cars and pedestrians that are missed or incorrectly classified by the RGB-only model. The nighttime RGB image is shown with a display-only gamma correction($\gamma = 0.4$).}

  \label{fig:qualitative}
  \vspace{-1.0em}
\end{figure}

The availability of such native LiDAR imagery opens new opportunities for applying image-based learning directly to LiDAR measurements. Previous studies have demonstrated the utility of LiDAR-generated range and reflectivity images for feature extraction~\cite{zhang2023lidar}, while more recent work has applied image colorization and super-resolution to LiDAR imagery encoding depth, reflectivity, and near-infrared information~\cite{ha2025enhancing}. 
Native LiDAR image representations have also been investigated for geometric estimation tasks, such as LiDAR–inertial odometry~\cite{pfreundschuh2024coin}.
These studies suggest that structured LiDAR imagery can serve as both a rich representation for learning-based perception and geometric estimation, rather than merely as an alternative representation of conventional point-cloud data.

However, despite the increasing availability and demonstrated utility of LiDAR image representations, their potential for dense semantic perception, particularly in camera–LiDAR fusion, remains underexplored.

Many existing fusion approaches operate on 3D point clouds or projected geometric representations, leaving the additional information available in native multi-channel LiDAR imagery underutilized~\cite{Gu2024CLFT}. Most camera-LiDAR fusion algorithms~\cite{vora2020pointpainting} still rely on raw 3D point coordinates as the LiDAR representation, discarding that additional channel information. Furthermore, point-cloud-based fusion methods require specialized preprocessing pipelines (voxelization, point set networks), which impose architectural complexity and prevent direct re-use of powerful vision-based models.

Meanwhile, Vision Transformers (ViTs)~\cite{dosovitskiy2020image} model global spatial dependencies via multi-head self-attention over image patches, providing strong capacity for dense, pixel-level prediction. Native LiDAR imagery therefore makes it possible to exploit image-pretrained vision architectures directly, without first converting LiDAR measurements into point-cloud representations.

Motivated by these observations, we propose TriLICT (Tri-LiDAR Image and Camera Transformer), a dual-stream transformer architecture that directly exploits native multi-channel LiDAR images together with RGB camera imagery for semantic segmentation. TriLICT constructs a three-channel LiDAR image by stacking three native LiDAR channels and processes it in parallel with an RGB image using two independent ViT encoders, allowing modality-specific representations to be learned within a common image-based framework. The resulting multi-scale features are hierarchically fused and decoded to produce pixel-level semantic predictions. We focus on the semantic segmentation of cars and pedestrians, two important foreground classes for autonomous driving perception. The main contributions of this study are as follows:

\begin{enumerate}[leftmargin=*, label=(\roman*)]

    \item We propose TriLICT, a dual-stream transformer that directly processes native multi-channel LiDAR imagery instead of only 3D-to-2D point-cloud projection, using independent ViT encoders for LiDAR and RGB imagery followed by hierarchical feature fusion for semantic segmentation.

    \item We train and systematically evaluate independent models for all $\binom{4}{3}=4$ three-channel LiDAR combinations with ViT-Base/16 and ViT-Large/16 backbones under daytime and nighttime conditions. With ViT-Base/16 and range, near-infrared, and reflectivity, TriLICT improves nighttime mIoU from $52.6\%$ to $74.0\%$ over the RGB-only baseline while maintaining comparable daytime performance (Fig.~\ref{fig:qualitative}). The best nighttime result, $74.8\%$ mIoU, is obtained with ViT-Base/16 and range, near-infrared, and signal.

    \item We benchmark TriLICT against two widely used camera-only segmentation networks, SegFormer-B2~\cite{xie2021segformer} and DeepLabV3+~\cite{chen2018deeplabv3plus}, trained and evaluated under the same protocol as our models. 
    At night, TriLICT with ViT-Base/16 reaches $74.0\%$ mIoU, about 18 points above the strongest camera-only network ($56.2\%$ for SegFormer-B2), with pedestrian IoU rising from $49.0\%$ to $74.3\%$. In daytime, TriLICT ($80.4\%$) is on par with DeepLabV3+ ($80.0\%$) and its RGB-only baseline ($80.6\%$) and 1.2 points below SegFormer-B2 ($81.6\%$), so the LiDAR stream adds robustness at night without costing daytime accuracy.

    \item We collect and publicly release a time-synchronized multimodal dataset spanning daytime and low-light nighttime conditions, comprising RGB imagery and four native Ouster LiDAR image channels, with pixel-level semantic annotations for background, cars, and pedestrians. The source code of TriLICT and the dataset are publicly available at \url{https://github.com/Jimmy-Zhang13798/TriLICT}.

\end{enumerate}

The remainder of this paper is organized as follows. Section~\ref{sec:related} reviews the related work, followed by a description of the proposed framework in Section~\ref{sec:methodology}. Sections~\ref{sec:experiment} and \ref{sec:results} present the experimental setup and analyze the results, respectively. Section~\ref{sec:conclusion} concludes the paper.

%% file: sec/02_RelatedWorks.tex
\section{Related Work}
\label{sec:related}

\subsection{Camera-LiDAR Fusion}
\label{subsec:rw_fusion}

Multi-modal fusion of camera and LiDAR has been studied extensively for autonomous driving perception. Depending on the fusion stage, approaches are broadly classified as early, late, or feature-level. Early fusion projects LiDAR points onto the image plane and combines the resulting LiDAR images with RGB at the input of the network~\cite{caltagirone2019lidar}. Late fusion processes each modality independently and merges predictions, limiting intermediate cross-modal interactions.

Feature-level fusion has become the dominant paradigm. PointPainting~\cite{vora2020pointpainting} enriches point clouds with semantic scores from a pretrained camera segmentation network, BEVFusion~\cite{liu2023bevfusion} unifies camera and LiDAR features in a shared bird's-eye-view (BEV) space, and MSeg3D~\cite{li2023mseg3d} combines intra-modal feature extraction with inter-modal feature fusion for point-wise 3D segmentation. Transformer-based methods further model long-range cross-modal dependencies, e.g.\ through the modality-agnostic query-based feature sampler of FUTR3D~\cite{Chen2022FUTR3D} or the soft-association decoder of TransFusion~\cite{bai2022transfusion}, which fuses object queries with image features. Our method follows the feature-level fusion paradigm, using dual independent ViT encoders on image-format inputs for both modalities without any 3D point-cloud operators.

\subsection{LiDAR Representation for 2D Neural Networks}
\label{subsec:rw_lidar_repr}

The choice of LiDAR representation determines which neural architectures can be applied. Raw 3D point sets are processed by PointNet~\cite{qi2017pointnet} with specialized set operators that preclude direct reuse of image-pretrained models, and voxelization~\cite{zhou2018voxelnet} enables 3D convolutions at a memory cost that scales cubically with spatial resolution. Spherical projection yields range images compatible with 2D networks, as in SqueezeSeg~\cite{wu2018squeezeseg}, RangeNet++~\cite{milioto2019rangenet}, and CENet~\cite{cheng2022cenet}, which segment full LiDAR scans efficiently with 2D convolutions.

\begin{figure}[h]
    \centering

    \includegraphics[width=0.98\columnwidth]{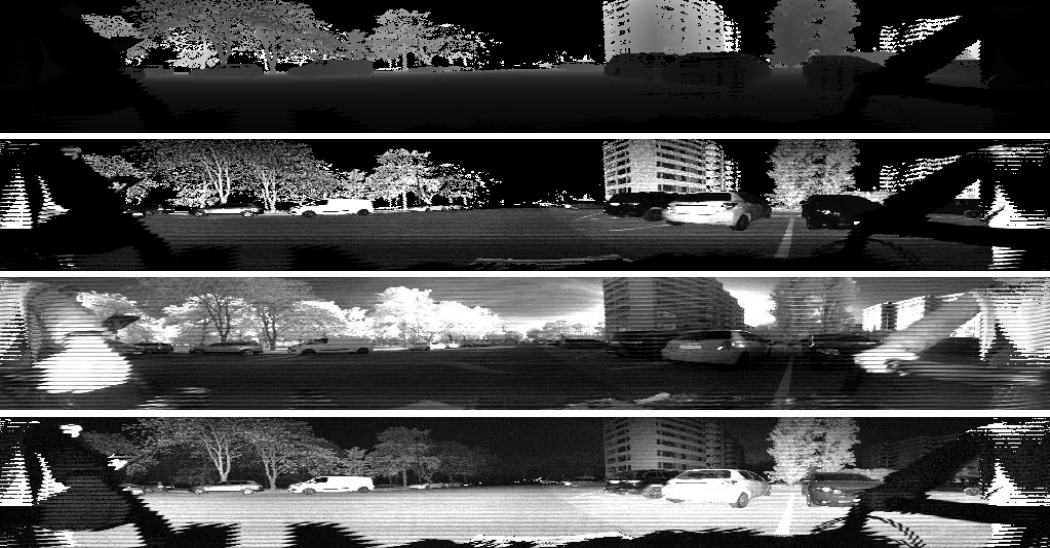}

    \caption{Different LiDAR image representations from our dataset. From top to bottom, the images correspond to the four native channels provided by the Ouster LiDAR: range, reflectivity, near-infrared, and signal.}
    \label{fig:different_lidar_images}
    \vspace{-0.5em}
\end{figure}

Another direction of LiDAR representation treats native LiDAR measurement directly as 2D images. Advanced LiDARs such as the Ouster natively output high-resolution images encoding range, near-infrared intensity, signal strength, and calibrated reflectivity, as shown in Fig.~\ref{fig:different_lidar_images}. Yu et al.~\cite{Yu2023} demonstrate that, with adequate image preprocessing, general-purpose detection and segmentation models built for camera images can be applied to such LiDAR images instead of processing the 3D point cloud, with all four evaluated detectors finding over 80\% of persons indoors and over 80\% of cars outdoors. Viswanath~et~al.~\cite{viswanath2024reflectivity} show that replacing raw intensity with calibrated reflectivity as a LiDAR input feature yields a 4\% mIoU improvement for off-road segmentation on Rellis-3D, and report potential benefits of calibrated intensity for urban segmentation on SemanticKITTI.

\subsection{Vision Transformers for Dense Prediction}
\label{subsec:rw_vit}

Vision Transformers~\cite{dosovitskiy2020image} model global spatial dependencies through patch-based self-attention, providing strong capacity for dense prediction tasks where both local detail and broad scene context must be considered. SETR~\cite{zheng2021setr} was among the first to apply a pure ViT encoder to semantic segmentation, framing it as a sequence-to-sequence prediction task and setting new benchmarks on ADE20K (50.28\% mIoU) and PASCAL Context (55.83\% mIoU). Dense Prediction Transformer (DPT)~\cite{Ranftl2021} introduced the Reassemble mechanism, which extracts intermediate ViT token sequences at four encoder depths, projects them into 2D spatial feature maps at corresponding scales, and fuses them via a convolutional decoder, achieving strong results on monocular depth estimation and semantic segmentation. SegFormer~\cite{xie2021segformer} pairs a hierarchical Mix Transformer encoder with a lightweight MLP decoder, reaching 51.0\% mIoU on ADE20K with significantly fewer parameters. Swin Transformer~\cite{liu2021swin} introduces shifted-window self-attention, reducing quadratic complexity while enabling cross-window interactions, and has become a widely adopted backbone for downstream perception tasks. Mask2Former~\cite{cheng2022mask2former} proposes masked attention that constrains cross-attention within predicted mask regions, achieving 57.7\% mIoU on ADE20K and establishing a unified framework for semantic, instance, and panoptic segmentation.

Our architecture is mainly built on DPT. We adopt the Reassemble mechanism and extend it to dual-modal inputs through two independent ViT encoder streams, one for RGB and one for the multi-channel LiDAR image.

\subsection{Multi-Modal Transformer Fusion}
\label{subsec:rw_multimodal}

Cross-modal attention and token-level fusion have been studied in both RGB-depth and camera-LiDAR settings. CMX~\cite{zhang2023cmx} proposes a cross-modal feature rectification module and a feature fusion module within a SegFormer-based framework for RGB-X segmentation, where X represents an arbitrary complementary modality—such as Depth, Thermal/Infrared, LiDAR, or Event camera. TokenFusion~\cite{wang2022tokenfusion} dynamically replaces uninformative tokens in one modality stream with tokens from the complementary modality, enabling adaptive cross-modal information exchange for multiple vision tasks. TransFuser~\cite{Prakash2021}  fuses image and LiDAR bird's-eye-view features at multiple resolutions through transformer self-attention within an end-to-end driving model that predicts waypoints. CLFT~\cite{Gu2024CLFT} extends the DPT architecture to dual-modal camera-LiDAR inputs, yet its LiDAR input is still derived from point-cloud coordinates, which are projected onto the camera plane and up-sampled into dense images.

TriLICT differs from the above works in two respects. First, rather than exchanging information between modalities inside the backbone, either by substituting tokens between modality streams~\cite{wang2022tokenfusion} or through a rectification mechanism between two parallel branches~\cite{zhang2023cmx}, it keeps two fully independent ViT encoders and fuses their features only in the decoder, so that each modality preserves its own representation throughout the backbone. Second, in contrast to CLFT~\cite{Gu2024CLFT} and other camera-LiDAR fusion methods whose LiDAR stream is fed with projections of point-cloud coordinates, TriLICT operates on native multi-channel Ouster images, maximizing the utilization of the reflectivity and intensity information that point-coordinate representations couldn't offer, while remaining directly compatible with image-pretrained encoders.

%% file: sec/03_Methodology.tex
\begin{figure*}[ht]
    \centering
    \color{red}
    \includegraphics[width=0.98\textwidth]{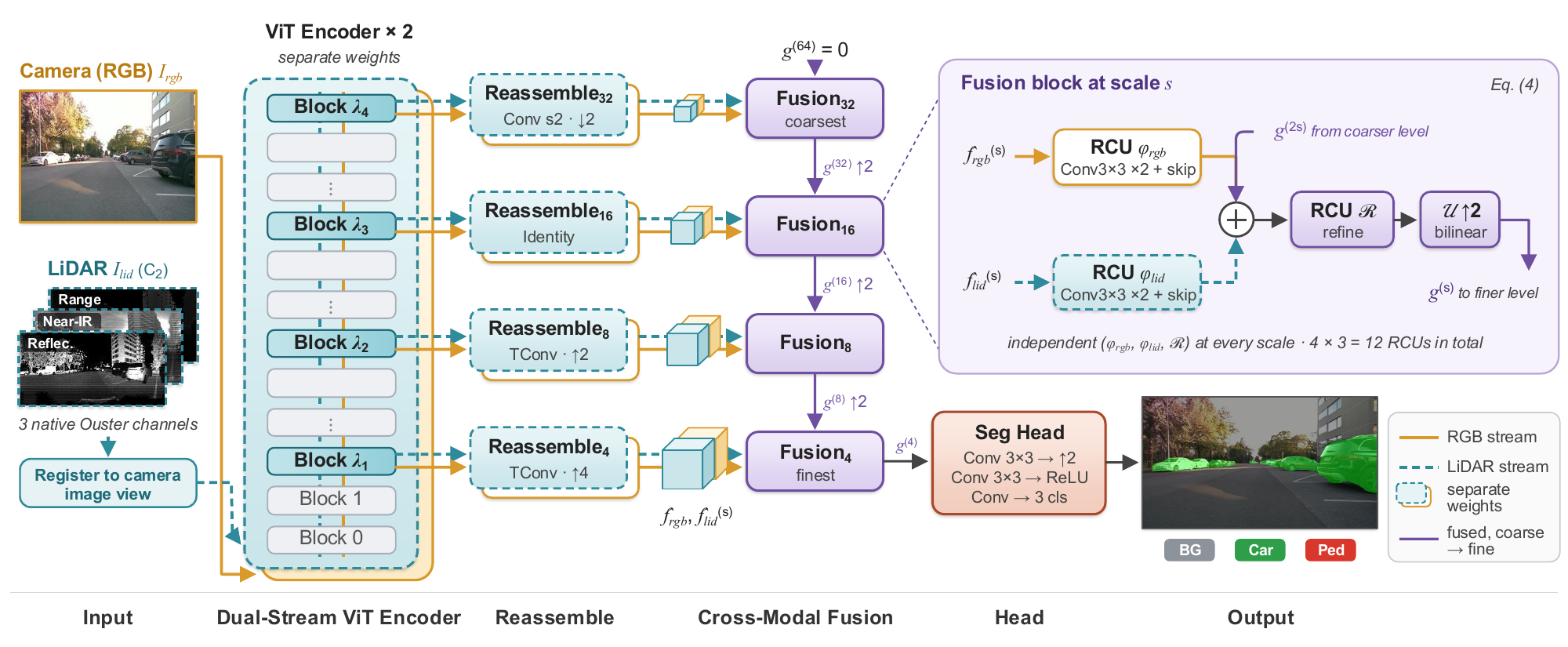}
    \caption{Overview of TriLICT. $\mathbf{I}_{\text{rgb}}$ and the camera-registered LiDAR image $\mathbf{I}_{\text{lid}}$ (here $\mathcal{C}_2$) are encoded by two ViTs with separate parameters (gold solid: RGB; teal dashed: LiDAR). Features from hooked blocks $\lambda_i \in \Lambda$ are reassembled per modality at $s \in \{32,16,8,4\}$ and fused coarse-to-fine by the Fusion decoder (inset: Eq.~(4)), followed by the segmentation head.}

    \label{fig:architecture}

\end{figure*}

\input{algs/trilict}

\section{Methodology}\label{sec:methodology}
\subsection{LiDAR Image Representation}\label{subsec:lidar_rep}

A key design decision in TriLICT is how to represent LiDAR data in a form that is directly compatible with a standard ViT backbone. The Ouster OS0-128 LiDAR outputs four independent measurement channels natively as grayscale images through its official driver. Detailed explanations are below.

\subsubsection{Native Ouster Image Format}

The Ouster OS0-128 driver produces four measurement channels directly as 2D images. Each channel is stored as a 16-bit grayscale image.

\subsubsection{Available Measurement Channels}
The LiDAR provides four grayscale images, each capturing a different physical property of the scene:

\paragraph*{\textbf{Range}} metric radial distance from the sensor origin to the target surface, stored in millimetres. Pixels without a valid return are encoded as zero.

\paragraph*{\textbf{Near-infrared (nearir)}} passive ambient near-infrared irradiance, integrated between laser pulses and therefore independent of the active return.

\paragraph*{\textbf{Signal}} active return intensity, proportional to surface reflectance and emitted laser power but also subject to the $1/r^{2}$\,($r$: distance) falloff of the returned pulse, so its value represents material properties and target distance.
\paragraph*{\textbf{Reflectivity (reflec)}} surface reflectance calibrated by the driver for range-dependent power falloff and beam-to-beam gain variation, and therefore a more faithful descriptor of material properties than the raw signal return.

The examples of these images can be found in Fig.~\ref{fig:different_lidar_images}.

\subsubsection{Channel Combination}

Using three of these channels encodes substantially richer scene semantics. We compose the three-channel LiDAR image ($\mathbf{I}_\text{lid}$) by selecting and stacking three of the four available channels. The use of three channels, rather than all four, is deliberate: it matches the input dimensionality of a standard RGB image, allowing the same ViT encoder to process both modalities without any architectural modification. Since four LiDAR channels are available, different three-channel combinations can be formed, enabling us to investigate which subset provides the most informative representation. The evaluated channel combinations are described below.

\begin{align}
    \mathcal{C}_1 &= (\text{range, nearir, signal}), \notag\\
    \mathcal{C}_2 &= (\text{range, nearir, reflec}), \notag\\
    \mathcal{C}_3 &= (\text{range, signal, reflec}), \notag\\
    \mathcal{C}_4 &= (\text{nearir, signal, reflec}).
    \label{eq:combos}
\end{align}

In our algorithm, an independent model is trained for each combination under identical conditions to identify which channel combination best supports our camera-LiDAR architecture.

\subsection{TriLICT Architecture}
\label{subsec:architecture}

We propose \textbf{TriLICT} (Tri-LiDAR Image and Camera Transformer), a dual-stream network that processes a three-channel LiDAR image $\mathbf{I}_\text{lid}$, composed of three distinct native channels, together with an RGB image $\mathbf{I}_\text{rgb}$ through four sequentially coupled stages: (1) a dual-stream ViT encoder that extracts multi-scale token features from each modality independently, (2) a reassemble module that converts token sequences into 2D spatial feature maps, (3) a cross-modal fusion decoder that integrates the two modalities hierarchically, and (4) a lightweight segmentation head. The complete forward pass is detailed in Algorithm~\ref{alg:trilict}, and the overall architecture is illustrated in Fig.~\ref{fig:architecture}.

\subsubsection{Dual-Stream ViT Encoder}
\label{subsubsec:encoder}

Both modality streams are processed by independent ViT encoders that share the same architecture but maintain separate parameters. Each encoder is initialized from ImageNet-pretrained weights, allowing the LiDAR encoder to specialize for LiDAR-domain features while benefiting from the rich spatial representations acquired during pretraining on natural images.

Given an input of size $H \times W$ with patch size $p$, the image is partitioned into $N=(H/p)(W/p)$ non-overlapping patches. Each patch is linearly projected into a $D$-dimensional token. A prepended classification token [\textsc{cls}] yields a sequence of $N+1$ tokens, which is processed through $L$ successive multi-head self-attention blocks. The learned positional embedding is used verbatim, with $N$ patch positions.

Forward hooks register callbacks at four designated intermediate blocks, indexed by the set $\Lambda$, capturing multi-scale representations at increasing semantic abstraction:
\begin{equation}
\Lambda = \begin{cases} \{2,\, 5,\, 8,\, 11\} & \text{ViT-Base} \\ \{5,\, 11,\, 17,\, 23\} & \text{ViT-Large} \end{cases}
\label{eq:hooks}
\end{equation}
Transformer blocks are 0-indexed, so the deepest hook is the final block of the encoder in both variants ($\lambda{=}11$ of 12 for ViT-Base, $\lambda{=}23$ of 24 for ViT-Large). 

We evaluate two backbone variants, ViT-Base ($p{=}16$, $D{=}768$, $L{=}12$) and ViT-Large ($p{=}16$, $D{=}1024$, $L{=}24$).

\subsubsection{Reassemble Module}
\label{subsubsec:reassemble}

The Reassemble module converts each hook's $(N{+}1) \times D$ token sequence into a 2D spatial feature map at unified channel width $\hat{D} = 256$, via three sequential operations.

\textit{Read}: the [\textsc{cls}] token is broadcast and concatenated with each of the $N$ patch tokens, forming, for every patch, a vector of dimension $2D$. An affine layer $W_p \in \mathbb{R}^{D \times 2D}$ with bias $b_p \in \mathbb{R}^{D}$, followed by GELU, projects each such vector back to $D$ dimensions, injecting global context into every spatial location.

\textit{Concat}: the $N$ processed tokens are rearranged from $(N, D)$ to $(D,\, H/p,\, W/p)$, recovering the 2D spatial layout.

\textit{Resample}: a $1{\times}1$ convolution maps channels from $D$ to $\hat{D}$. A scale-specific operator $\mathcal{T}_s$ then brings the spatial resolution to $\tfrac{H}{s} \times \tfrac{W}{s}$:
\begin{equation}
\mathcal{T}_s = \begin{cases}
\text{TConv}_{4\times4}^{\,\text{stride }4} & s = 4 \quad (4{\times}\text{ upsampling})\\
\text{TConv}_{2\times2}^{\,\text{stride }2} & s = 8 \quad (2{\times}\text{ upsampling})\\
\mathrm{Id} & s = 16 \\
\text{Conv}_{2\times2}^{\,\text{stride }2} & s = 32 \quad (2{\times}\text{ downsampling})
\end{cases}
\label{eq:Ts}
\end{equation}
all at constant channel width $\hat{D}$. Each hook index $\lambda \in \Lambda$ is paired with exactly one scale $s \in \mathcal{S} = \{4, 8, 16, 32\}$ in ascending order, so that shallower blocks supply higher-resolution features, following the design of DPT~\cite{Ranftl2021}.

Separate Reassemble branches are maintained for each modality, producing $\hat{\mathbf{f}}^{(s)}_\text{rgb}$ and $\hat{\mathbf{f}}^{(s)}_\text{lid} \in \mathbb{R}^{\hat{D} \times \frac{H}{s} \times \frac{W}{s}}$ for $s \in \mathcal{S}$. For readability, the batch dimension $B$ is omitted from all tensor shapes in this section; it is retained explicitly in Algorithm~\ref{alg:trilict}.

\subsubsection{Cross-Modal Fusion}
\label{subsubsec:fusion}

The Fusion decoder integrates RGB and LiDAR feature maps in a coarse-to-fine (top-down) pass from the coarsest level ($s{=}32$) to the finest ($s{=}4$), in the sense used in the FPN literature, where ``bottom-up'' refers to the encoder direction. At each scale $s$:
\begin{equation}
    \mathbf{g}^{(s)} = \mathcal{U}\!\left(\mathcal{R}\!\left(\phi_\text{rgb}\!\left(\hat{\mathbf{f}}^{(s)}_\text{rgb}\right) + \phi_\text{lid}\!\left(\hat{\mathbf{f}}^{(s)}_\text{lid}\right) + \mathbf{g}^{(2s)}\right)\right),
    \label{eq:fusion}
\end{equation}
where $\phi_\text{rgb}$ and $\phi_\text{lid}$ are modality-specific Residual Convolutional Units (RCUs), $\mathcal{U}(\cdot)$ denotes bilinear $2{\times}$ upsampling applied to the fused output before passing it to the next finer scale, and $\mathbf{g}^{(64)} = \mathbf{0}$ (zero tensor at resolution $\tfrac{H}{32} \times \tfrac{W}{32}$) at the coarsest boundary. Each RCU applies two $3{\times}3$ convolutions with ReLU activations and an identity skip connection at constant channel width $\hat{D}$.

\textit{Parameter sharing.} every scale $s \in \mathcal{S}$ instantiates its own independent triple $(\phi_\text{rgb}^{(s)},\, \phi_\text{lid}^{(s)},\, \mathcal{R}^{(s)})$, and the superscript is dropped in Eq.\eqref{eq:fusion} only for brevity. The decoder therefore holds $4 \times 3 = 12$ RCUs, each with two $3{\times}3$ convolutions at $\hat{D}{=}256$ channels, i.e.\ $12 \times 2 \times (256^2 \times 9 + 256) \approx 14.2$M parameters, independent of the backbone choice.

\textit{Resolution convention.} Because $\mathcal{U}$ is applied at the end of each level, the superscript of $\mathbf{g}^{(s)}$ labels the level that produced it, and its resolution is $\tfrac{2H}{s} \times \tfrac{2W}{s}$. Hence $\mathbf{g}^{(2s)}$ is spatially aligned with $\hat{\mathbf{f}}^{(s)}$, and the sum in Eq.~\eqref{eq:fusion} is dimensionally consistent at every level.

Since $\mathbf{I}_\text{lid}$ consists of native LiDAR images registered to the camera view as elucidated in Section~\ref{subsubsec:registration}, $\hat{\mathbf{f}}^{(s)}_\text{rgb}$ and $\hat{\mathbf{f}}^{(s)}_\text{lid}$ are spatially aligned at every scale, so the element-wise sum in Eq.~\eqref{eq:fusion} combines features of the same scene location.

\subsubsection{Segmentation Head}
\label{subsubsec:head}

The finest-scale output $\mathbf{g}^{(4)} \in \mathbb{R}^{256 \times \frac{H}{2} \times \frac{W}{2}}$ is passed through a three-convolution head (Eq.~\eqref{eq:head}) that restores the input resolution and yields per-class logits $\hat{\mathbf{Y}} \in \mathbb{R}^{C \times H \times W}$ with $C = 3$. Compared with the original DPT head~\cite{Ranftl2021}, it omits batch normalization and the intermediate ReLU, which lowers the decoder's memory footprint; the preceding RCU stack already supplies the nonlinear capacity, and all configurations use this identical head.

\begin{equation}
\begin{aligned}
    \mathbf{h} &= \text{ReLU}\big(\text{Conv}_{3\times3}^{128\to32}
    \big(\mathcal{U}(\text{Conv}_{3\times3}^{256\to128}(\mathbf{g}^{(4)}))\big)\big),\\
    \hat{\mathbf{Y}} &= \text{Conv}_{1\times1}^{32\to C}(\mathbf{h}),
\end{aligned}
\label{eq:head}
\end{equation}

\subsection{Training Configuration}
\label{subsec:training}

\subsubsection{Loss Function}
We minimize a class-weighted cross-entropy loss to compensate for the severe foreground/background imbalance in outdoor driving scenes:
\begin{equation}
    \mathcal{L} = \frac{-\displaystyle\sum_{i \in \Omega} \sum_{c=0}^{C-1} w_c \, y_{i,c} \log \hat{p}_{i,c}}
                       {\displaystyle\sum_{i \in \Omega} \sum_{c=0}^{C-1} w_c \, y_{i,c}},
    \label{eq:loss}
\end{equation}
where $\Omega$ is the set of all pixels in the mini-batch, $|\Omega| = B \cdot H \cdot W$, $y_{i,c}$ is the ground-truth one-hot label for pixel $i$ and class $c$, and $\hat{p}_{i,c}$ is the predicted softmax probability. The denominator is the sum of the weights of the ground-truth classes present in the batch, i.e.\ Eq.\eqref{eq:loss} is a weight-normalized mean rather than a raw sum.

Class weights are set to $w_0{=}1$ (background), $w_1{=}4$ (cars), and $w_2{=}10$ (pedestrians), i.e.\ inversely ordered with respect to annotation frequency: background dominates the pixel count, cars are the more frequent foreground class, and pedestrians the rarest.

\subsubsection{Modality Configurations}

Two network variants are trained:
\begin{itemize}[noitemsep]
    \item \textbf{Dual-fusion (proposed)}: both encoders are active and their features are jointly processed in Eq.\eqref{eq:fusion}.

    \item \textbf{RGB-only (baseline)}: the LiDAR encoder is never executed and the entire operator $\phi_\text{lid}$ is \emph{removed} from Eq.\eqref{eq:fusion}, which reduces to
    \begin{equation}
        \mathbf{g}^{(s)} = \mathcal{U}\!\left(\mathcal{R}\!\left(\phi_\text{rgb}\!\left(\hat{\mathbf{f}}^{(s)}_\text{rgb}\right) + \mathbf{g}^{(2s)}\right)\right).
        \label{eq:fusion_rgb}
    \end{equation}

\end{itemize}

%% file: algs/trilict.tex
\begin{algorithm*}[!t]
\small
\SetAlgoLined
\DontPrintSemicolon
\SetCommentSty{normalfont}
\SetKwInOut{KwPar}{Parameters}
\caption{TriLICT Forward Pass}
\label{alg:trilict}
\KwIn{$\mathbf{I}_\text{rgb}\!\in\!\mathbb{R}^{B\times3\times H\times W}$ (RGB image);\enspace
      $\mathbf{I}_\text{lid}\!\in\!\mathbb{R}^{B\times3\times H\times W}$ (LiDAR image)}
\KwOut{$\hat{\mathbf{Y}}\in\mathbb{R}^{B\times C\times H\times W}$,\enspace per-pixel class logits ($C{=}3$)}
\KwPar{$p{=}16$,\; $D{=}768$,\; $\hat{D}{=}256$,\; $N{=}(H/p)(W/p)$;\enspace
       $\Lambda{=}\{2,5,8,11\}$ (0-indexed hook blocks),\; $\mathcal{S}{=}\{4,8,16,32\}$ (scales);}
\tcp*[f]{\emph{Parameters instantiated for ViT-Base; ViT-Large uses $D{=}1024$, $\Lambda{=}\{5,11,17,23\}$. $H,W$ denote image height and width only.}}
\BlankLine

\tcp{\textbf{Stage 1}\quad Dual-Stream ViT Encoding with Forward Hooks}
$\bigl\{\mathbf{t}^{(\lambda)}_\text{rgb}\bigr\}_{\lambda\in\Lambda}
   \leftarrow\mathrm{ViT}_\text{rgb}(\mathbf{I}_\text{rgb})$,\quad
  $\mathbf{t}^{(\lambda)}_\text{rgb}\!\in\!\mathbb{R}^{B\times(N+1)\times D}$\;
$\bigl\{\mathbf{t}^{(\lambda)}_\text{lid}\bigr\}_{\lambda\in\Lambda}
   \leftarrow\mathrm{ViT}_\text{lid}(\mathbf{I}_\text{lid})$,\quad
  $\mathbf{t}^{(\lambda)}_\text{lid}\!\in\!\mathbb{R}^{B\times(N+1)\times D}$\;
\BlankLine

\tcp{\textbf{Stage 2}\quad Reassemble: Token Sequences $\to$ 2D Multi-Scale Feature Maps}
\For{$s\in\{4,8,16,32\}$ with paired $\lambda\in\Lambda$}{
  \tcp{\textbf{Read:}\enspace broadcast [\textsc{cls}] token; concat with each of $N$ patch tokens; \textsc{gelu}-project to $D$ dims}
  $\mathbf{e}^{(\lambda)}_m\!\leftarrow\!
   \text{GELU}\!\bigl([\,\mathbf{t}^{(\lambda)}_{m,\,1:}
   \;\|\;\mathbf{1}_N\mathbf{t}^{(\lambda)\top}_{m,\,0}]\,W_p^{\top}+b_p\bigr)
   \in\mathbb{R}^{B\times N\times D},\;\; W_p\!\in\!\mathbb{R}^{D\times2D},\, b_p\!\in\!\mathbb{R}^{D},\;\, m\in\{\text{rgb},\text{lid}\}$\;
  \tcp{\textbf{Concat:}\enspace rearrange flat token sequence to 2D spatial grid}
  $\mathbf{M}^{(\lambda)}_m\!\leftarrow\!
   \mathrm{Rearrange}\!\bigl(\mathbf{e}^{(\lambda)}_m:\;B,N,D\to B,D,\tfrac{H}{p},\tfrac{W}{p}\bigr)
   \in\mathbb{R}^{B\times D\times\frac{H}{p}\times\frac{W}{p}}$\;
  \tcp{\textbf{Resample:}\enspace $1{\times}1$ conv ($D\!\to\!\hat{D}$); scale-specific spatial op to resolution $H/s\!\times\!W/s$}
  $\hat{\mathbf{f}}^{(s)}_\text{rgb}\!\leftarrow\!
   \mathcal{T}_{s}\!\bigl(\text{Conv}^{D\to\hat{D}}_{1\times1}(\mathbf{M}^{(\lambda)}_\text{rgb})\bigr)
   \in\mathbb{R}^{B\times\hat{D}\times\frac{H}{s}\times\frac{W}{s}}$
  $\hat{\mathbf{f}}^{(s)}_\text{lid}\!\leftarrow\!
   \mathcal{T}_{s}\!\bigl(\text{Conv}^{D\to\hat{D}}_{1\times1}(\mathbf{M}^{(\lambda)}_\text{lid})\bigr)
   \in\mathbb{R}^{B\times\hat{D}\times\frac{H}{s}\times\frac{W}{s}}$
}
\BlankLine

\tcp{\textbf{Stage 3}\quad Hierarchical Cross-Modal Fusion (coarse $s{=}32\;\to$ fine $s{=}4$)}
$\mathbf{g}^{(64)}\leftarrow\mathbf{0}
 \in\mathbb{R}^{B\times\hat{D}\times\frac{H}{32}\times\frac{W}{32}}$\;
\For{$s\in\{32,16,8,4\}$}{
  $\mathbf{g}^{(s)}\leftarrow
   \mathcal{U}\!\Bigl(\mathcal{R}\!\bigl(
     \phi_\text{rgb}(\hat{\mathbf{f}}^{(s)}_\text{rgb})
     +\phi_\text{lid}(\hat{\mathbf{f}}^{(s)}_\text{lid})
     +\mathbf{g}^{(2s)}\bigr)\Bigr)$
  \tcp*{per-scale RCUs $\phi^{(s)}_\text{rgb},\phi^{(s)}_\text{lid}$;\enspace $\mathcal{R}^{(s)}$: RCU shared across modalities;\enspace $\mathcal{U}$: bilinear ${\times}2$ \enspace
}
}
\BlankLine

\tcp{\textbf{Stage 4}\quad Segmentation Head:\enspace
  $\mathbf{g}^{(4)}\!\in\!\mathbb{R}^{B\times\hat{D}\times\frac{H}{2}\times\frac{W}{2}}
  \;\longrightarrow\;\hat{\mathbf{Y}}\!\in\!\mathbb{R}^{B\times C\times H\times W}$}
$\mathbf{h}_1\leftarrow\text{Conv}_{3\times3}^{256\to128}(\mathbf{g}^{(4)})$
\tcp*{$\in\mathbb{R}^{B\times128\times\frac{H}{2}\times\frac{W}{2}}$}
$\mathbf{h}_2\leftarrow\mathcal{U}(\mathbf{h}_1)$
\tcp*{$\in\mathbb{R}^{B\times128\times H\times W}$}
$\mathbf{h}_3\leftarrow\text{ReLU}\!\left(\text{Conv}_{3\times3}^{128\to32}(\mathbf{h}_2)\right)$
\tcp*{$\in\mathbb{R}^{B\times32\times H\times W}$}
$\hat{\mathbf{Y}}\leftarrow\text{Conv}_{1\times1}^{32\to C}(\mathbf{h}_3)$
\tcp*{$\in\mathbb{R}^{B\times C\times H\times W}$}
\BlankLine
\Return $\hat{\mathbf{Y}}$
\end{algorithm*}

%% file: sec/04_Experiment.tex
\begin{figure}[h]
    \centering
    \color{red}
    \includegraphics[width=0.98\columnwidth]{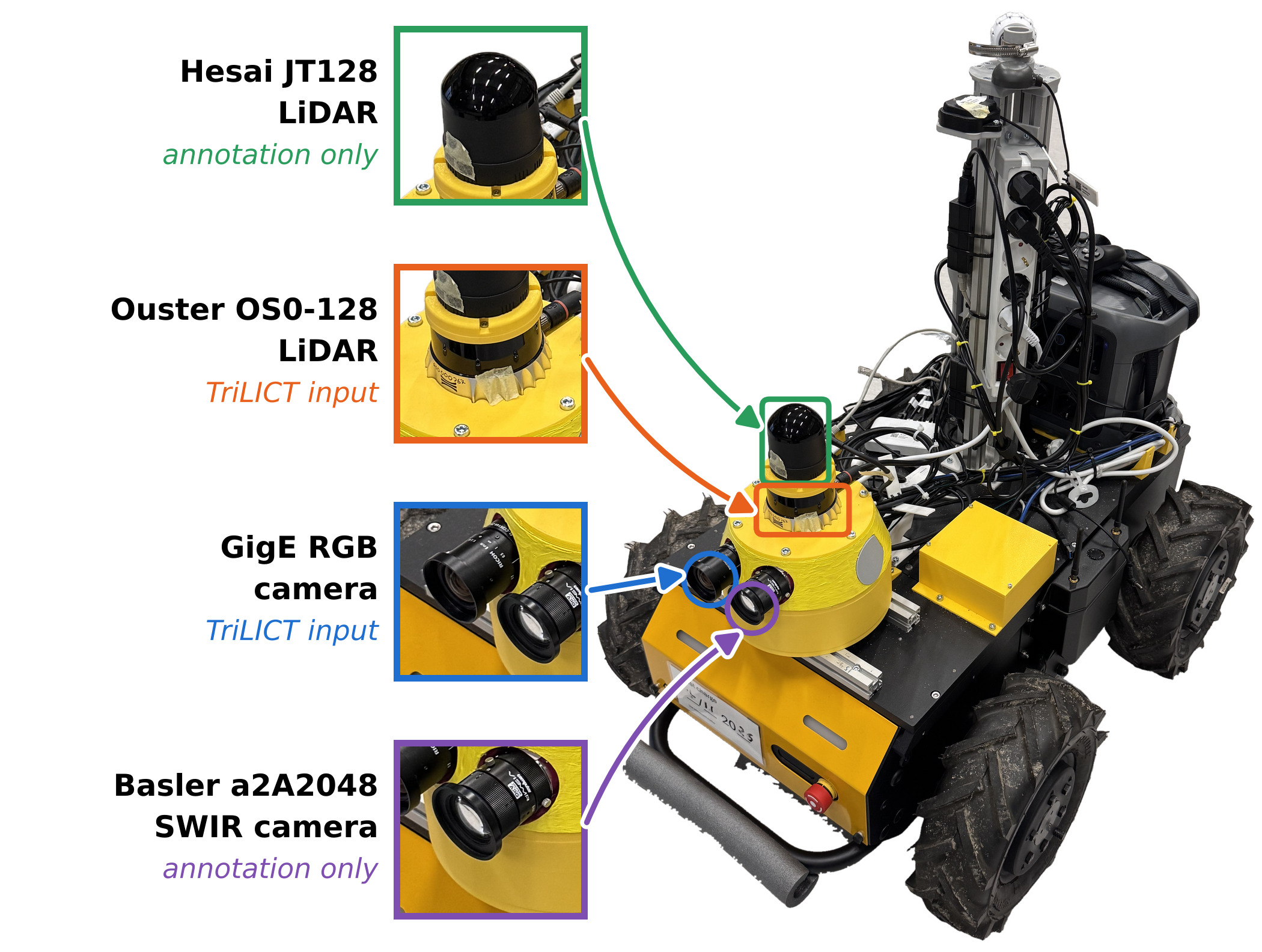}
    \caption{Sensors and mobile robot setup.}
    \label{fig:platform}
\end{figure}

\section{Experiment}\label{sec:experiment}

\subsection{Dataset}
\label{subsec:dataset}

\subsubsection{Collection}
\label{subsubsec:collection}

We conduct our experiments on a custom dataset collected with a Husky A200 mobile platform equipped with an Ouster OS0-128 LiDAR and a front-facing RGB camera: GigE, as shown in Fig.~\ref{fig:platform}. Data were recorded in two scenarios, daytime and nighttime. In both scenarios, the camera and the LiDAR are time-synchronized and extrinsically calibrated.

\subsubsection{Annotation}
\label{subsubsec:annotation}

Pixel-level semantic annotations are defined in the camera view for three classes: \textit{background} (class 0), \textit{cars} (class 1), and \textit{pedestrians} (class 2).

\paragraph{Daytime annotation}
Preliminary labels were first generated automatically on the RGB images using SAM3~\cite{carion2026sam3segmentconcepts}. They were then imported into Roboflow~\cite{dwyer2022roboflow}, where every frame was manually reviewed and refined.

\paragraph{Nighttime annotation}
At night, the RGB images carry very little usable information: most of the scene appears almost black, and even after low-light enhancement, objects cannot be annotated reliably from the camera image alone. We therefore annotate the nighttime data using several sensing modalities. Labels are drawn on the Ouster OS0-128 data, in both the native LiDAR images and the 3D point cloud. In addition, two further sensors were mounted on the platform: a Basler a2A2048 short-wave infrared (SWIR) camera and a Hesai JT128 LiDAR. All modalities are calibrated, combined to produce automatic, model-assisted pre-annotations, and every frame is then manually reviewed. Finally, all annotations are transferred into the RGB camera view using the calibrated extrinsics, which yields pixel-level semantic segmentation ground truth for the camera images. The SWIR camera and the Hesai LiDAR are used only for annotation and are not inputs to TriLICT.

Table~\ref{tab:dataset_stats} summarizes the training, validation, and test splits of both scenarios.

\begin{table}[h]
\centering
\captionsetup{font=footnotesize}
\caption{Dataset split statistics.}
\label{tab:dataset_stats}
\resizebox{0.98\linewidth}{!}{
\begin{tabular}{lcccc}
\toprule
\textbf{Scenario} & \textbf{Train} & \textbf{Val} & \textbf{Test} & \textbf{Total} \\
\midrule
Daytime   & 2{,}071 & 444 & 443 & 2{,}958 \\
Nighttime & 3{,}150 & 636 & 700 & 4{,}486 \\
\bottomrule
\end{tabular}
}
\end{table}

\subsubsection{Registration to the Camera View}
\label{subsubsec:registration}

The native Ouster images form a $360^\circ$ panorama, whereas the camera observes only a limited perspective field of view. Since the camera and the LiDAR are rigidly mounted and calibrated, we register the native LiDAR images to the camera view in a one-off offline preprocessing step, so that every LiDAR pixel fed to TriLICT corresponds to the same viewing direction as the RGB pixel at the same location.

\subsection{Implementation Pipeline}
\label{subsec:impl}

To investigate how different parameter sizes affect the performance, we experiment with two distinct backbone configurations. The two backbone configurations evaluated in this work are summarized in Table~\ref{tab:backbones}.

\begin{table}[h]
\centering
\captionsetup{font=footnotesize}
\caption{ViT backbone configurations used in the experiments.}
\label{tab:backbones}
\resizebox{0.98\linewidth}{!}{
\begin{tabular}{lcccc} 
\toprule
\textbf{Backbone} & $p$ & $D$ & $L$ & Hooks $\Lambda$ (0-indexed) \\ 
\midrule
ViT-Base  & 16 & 768  & 12 & $\{2,5,8,11\}$  \\
ViT-Large & 16 & 1024 & 24 & $\{5,11,17,23\}$ \\
\bottomrule
\end{tabular}
}
\end{table}

The daytime and nighttime scenarios are trained independently, which gives $2 \times 2 \times 4 = 16$ dual-fusion models (scenario $\times$ backbone $\times$ LiDAR combination), each trained once with the same random seed. The RGB-only baseline does not depend on the LiDAR combination; each of its $2 \times 2 = 4$ configurations is trained with four random seeds (0--3), so that it can be reported as mean $\pm$ standard deviation and the differences to fusion can be judged against the run-to-run variance. All 32 runs use the identical training recipe.

\subsection{Benchmark}
\label{sec:benchmark}

Besides the RGB-only method trained together with TriLICT, we also implement two separate RGB segmentation algorithms for comparison: SegFormer-B2~\cite{xie2021segformer}, a hierarchical Transformer with an all-MLP decoder, and DeepLabV3+~\cite{chen2018deeplabv3plus} with a ResNet-101 encoder~\cite{he2015deepresiduallearningimage}. For a fair comparison, both use ImageNet pre-trained encoders and the same data, augmentation, loss, optimiser, schedule, early stopping and evaluation code as TriLICT; only the network is replaced.

We further compare with CLFT~\cite{Gu2024CLFT}, a camera--LiDAR fusion transformer that extends the DPT architecture to camera and LiDAR inputs. Following CLFT, its LiDAR input consists of the $X$, $Y$, $Z$ coordinates of the point cloud projected onto the camera plane and up-sampled into a three-channel image. It uses the same data splits, test labels and evaluation code as all other methods.

\subsection{Evaluation Metrics}
\label{subsec:metrics}

All metrics are computed from the pixel-level true positives $\text{TP}_c$, false positives $\text{FP}_c$, and false negatives $\text{FN}_c$ of class $c$, accumulated over all frames of a split. The primary metric is the per-class Intersection over Union and its mean over the two foreground classes, with background excluded:
\begin{equation*}
    \text{IoU}_c = \frac{\text{TP}_c}{\text{TP}_c + \text{FP}_c + \text{FN}_c}, \qquad
    \text{mIoU} = \tfrac{1}{2}\left(\text{IoU}_{\text{cars}} + \text{IoU}_{\text{ped}}\right).
\end{equation*}
The checkpoint with the highest validation mIoU is retained for testing. For the detailed evaluation (Section~\ref{subsec:detailed}), we additionally report per-class precision $\text{TP}_c/(\text{TP}_c+\text{FP}_c)$, recall $\text{TP}_c/(\text{TP}_c+\text{FN}_c)$, their harmonic mean F1, and the average precision AP, i.e.\ the area under the precision--recall curve obtained by thresholding the softmax score of class $c$. Over all three classes including background, we report pixel accuracy (PA), mean per-class accuracy (mAcc), and frequency-weighted IoU (FWIoU), in which each $\text{IoU}_c$ is weighted by the share of ground-truth pixels of class $c$.

\subsection{Hardware and Software Setup}
\label{subsec:hardware}

The model training and evaluation are conducted on a workstation equipped with two NVIDIA GeForce RTX 3090 GPUs (24\,GB VRAM each). The core software libraries comprise Python 3.9.20, PyTorch 2.1.0 (CUDA 11.8, cuDNN 8.7.0), and timm 0.9.8.

%% file: sec/05_Results.tex
\section{Results}
\label{sec:results}

\subsection{Fusion vs.\ RGB-Only Baseline}
\label{subsec:modality}

Table~\ref{tab:modality} compares the RGB-only baseline, reported as mean $\pm$ standard deviation over four seeds (Section~\ref{subsec:impl}), with dual-fusion TriLICT using $\mathcal{C}_2$ \{range, nearir, reflec\} on both test splits.

\begin{table}[!t]
\centering
\captionsetup{font=footnotesize}
\caption{Modality comparison. IoU (\%) on the daytime and nighttime test splits. RGB-only: mean $\pm$ standard deviation over four seeds; dual fusion: $\mathcal{C}_2$ \{range, nearir, reflec\}. $\Delta$: mIoU of dual fusion minus the RGB-only mean.}
\label{tab:modality}
\resizebox{\linewidth}{!}{
\begin{tabular}{lllcccc}
\toprule
\textbf{Scenario} & \textbf{Backbone} & \textbf{Modality} & Cars IoU & People IoU & mIoU & $\Delta$ \\
\midrule
\multirow{4}{*}{Daytime}
  & \multirow{2}{*}{ViT-Base/16}  & RGB-only           & 94.3 $\pm$ 0.1 & 66.9 $\pm$ 0.8 & 80.6 $\pm$ 0.4 & -- \\
  &                               & Dual-fusion (ours) & 94.2 & 66.7 & 80.4 & $-0.2$ \\
  & \multirow{2}{*}{ViT-Large/16} & RGB-only           & 94.2 $\pm$ 0.1 & 69.0 $\pm$ 0.6 & 81.6 $\pm$ 0.3 & -- \\
  &                               & Dual-fusion (ours) & 94.2 & 68.7 & 81.5 & $-0.1$ \\
\midrule
\multirow{4}{*}{Nighttime}
  & \multirow{2}{*}{ViT-Base/16}  & RGB-only           & 63.5 $\pm$ 0.4 & 41.6 $\pm$ 2.0 & 52.6 $\pm$ 0.9 & -- \\
  &                               & Dual-fusion (ours) & \textbf{73.7} & \textbf{74.3} & \textbf{74.0} & $\mathbf{+21.4}$ \\
  & \multirow{2}{*}{ViT-Large/16} & RGB-only           & 63.1 $\pm$ 0.5 & 38.6 $\pm$ 4.6 & 50.9 $\pm$ 2.5 & -- \\
  &                               & Dual-fusion (ours) & \textbf{74.1} & \textbf{74.0} & \textbf{74.1} & $\mathbf{+23.2}$ \\
\bottomrule
\end{tabular}
}
\end{table}

In daytime, dual fusion and RGB-only perform equivalently: TriLICT reaches 80.4\% (ViT-Base/16) and 81.5\% (ViT-Large/16) mIoU against 80.6\,$\pm$\,0.4\% and 81.6\,$\pm$\,0.3\% for the baseline. The reason is that under good illumination the RGB image already carries sufficient cues.

At night the RGB-only baseline falls to 52.6\,$\pm$\,0.9\% (ViT-Base/16) and 50.9\,$\pm$\,2.5\% (ViT-Large/16) mIoU, whereas TriLICT keeps 74.0\% and 74.1\%, an absolute gain of 21.4~pp and 23.2~pp. Pedestrians benefit most: People~IoU rises by 32.7--35.4~pp (41.6\%~$\to$~74.3\% and 38.6\%~$\to$~74.0\%), while Cars~IoU improves by 10.2--11.0~pp, as large and often partly lit cars remain partly visible in the dark camera image. The nighttime RGB-only models are also unstable: People~IoU varies by up to $\pm$4.6~pp across seeds, and their validation mIoU peaks within the first 7--35 epochs before declining, whereas the fusion models reach a stable validation plateau. Fig.~\ref{fig:qualitative} illustrates this failure mode: on the almost black camera frame, the RGB-only model misses the pedestrian and predicts a non-existent car, whereas TriLICT segments the pedestrian and both cars.

\subsection{Other Baseline Comparison}
\label{subsec:benchmark_results}

Table~\ref{tab:baseline_comparison} places TriLICT next to the camera-only networks of Section~\ref{sec:benchmark}, all trained and evaluated under the identical protocol. It also includes the camera--LiDAR fusion method CLFT-Base~\cite{Gu2024CLFT} (Section~\ref{sec:benchmark}).

\begin{table}[!t]
\centering
\captionsetup{font=footnotesize}
\caption{Comparison with other segmentation networks trained under the same protocol. IoU (\%) on the test splits. C: camera, L: LiDAR. RGB-only: mean over four seeds; TriLICT: $\mathcal{C}_2$ \{range, nearir, reflec\}. CLFT-Base: camera--LiDAR fusion with the projected point coordinates ($X$, $Y$, $Z$) as LiDAR input. Bold: best per column; underlined: second best (ranked on unrounded values).}
\label{tab:baseline_comparison}
\resizebox{\linewidth}{!}{
\begin{tabular}{llcccccc}
\toprule
\multirow{2}{*}{\textbf{Method}} & \multirow{2}{*}{\textbf{Input}} & \multicolumn{3}{c}{\textbf{Daytime}} & \multicolumn{3}{c}{\textbf{Nighttime}} \\
\cmidrule(lr){3-5}\cmidrule(lr){6-8}
 & & Cars & People & mIoU & Cars & People & mIoU \\
\midrule

RGB-only (ViT-Base/16)  & C   & 94.3 & \underline{66.9} & \underline{80.6} & 63.5 & 41.6 & 52.6 \\
SegFormer-B2            & C   & \textbf{94.9} & \textbf{68.3} & \textbf{81.6} & 63.4 & 49.0 & 56.2 \\
DeepLabV3+ (ResNet-101) & C   & \underline{94.7} & 65.3 & 80.0 & 60.5 & 34.1 & 47.3 \\
\midrule

CLFT-Base & C+L & 93.2 & 64.8 & 79.0 & \underline{71.7} & \underline{70.4} & \underline{71.0} \\
TriLICT (ViT-Base/16)   & C+L & 94.2 & 66.7 & 80.4 & \textbf{73.7} & \textbf{74.3} & \textbf{74.0} \\
\bottomrule
\end{tabular}
}
\end{table}

\begin{figure*}[!t]
  \centering
  \includegraphics[width=\textwidth]{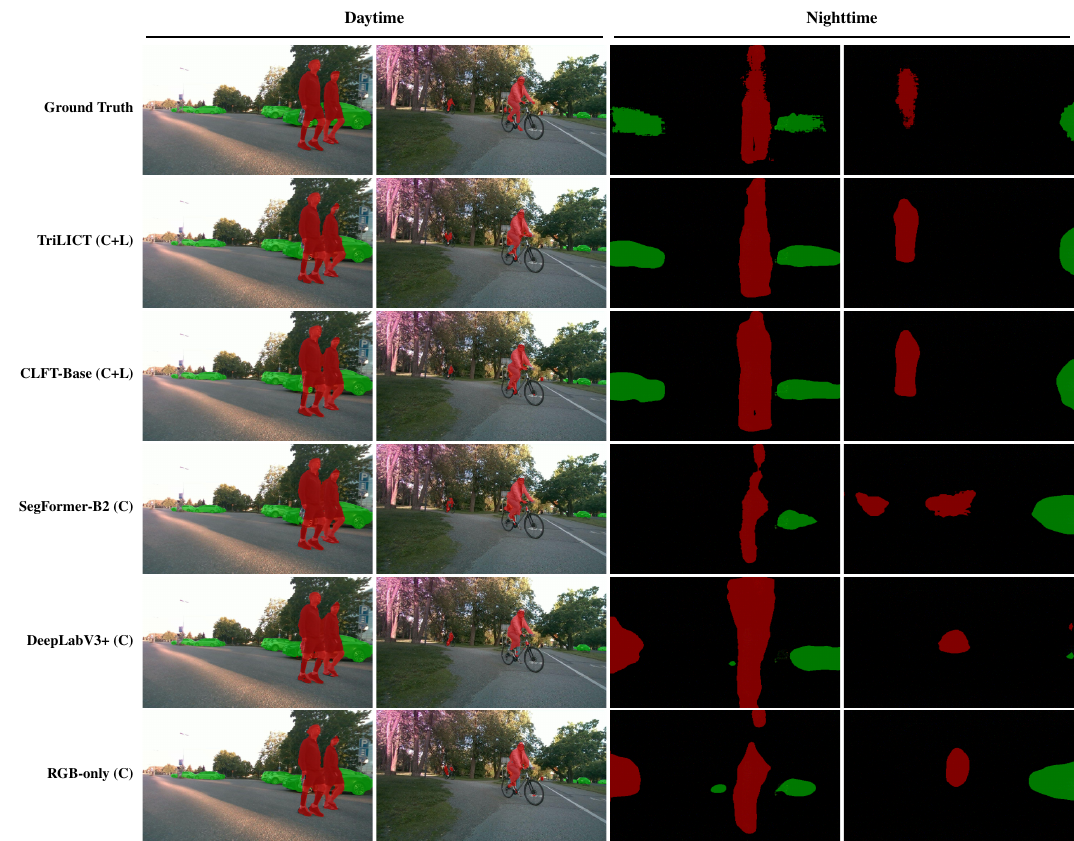}
  \caption{Qualitative comparison of the methods in Table~\ref{tab:baseline_comparison} on the daytime (left two columns) and nighttime (right two columns) test splits. Predictions are overlaid on the camera image; cars are shown in green and people in red. C: camera, L: LiDAR. The nighttime camera images are shown with a display-only gamma correction ($\gamma=0.4$) and are cropped to the aspect ratio of the daytime images.}
  \label{fig:method_comparison}
\end{figure*}

In daytime, TriLICT (80.4\%) is on par with the RGB-only baseline (80.6\%) and DeepLabV3+ (80.0\%) and 1.2~pp below SegFormer-B2 (81.6\%), mainly through a lower People~IoU (66.7\% vs.\ 68.3\%).

At night, every camera-only network degrades sharply, most strongly on pedestrians. The best of them, SegFormer-B2, reaches 56.2\% mIoU and DeepLabV3+ only 47.3\%, so the camera-only architectures differ by up to 8.9~pp among themselves; yet TriLICT (74.0\%) exceeds them by 17.8~pp and 26.7~pp, and its People~IoU (74.3\%) is 25.3~pp above the best camera-only result (49.0\%). A stronger camera-only architecture therefore cannot compensate for the visual information missing at night; the advantage of TriLICT comes from the LiDAR modality.

CLFT-Base, which also fuses camera and LiDAR, confirms the value of the LiDAR modality at night: its 71.0\% mIoU is 14.8~pp above SegFormer-B2, and its People~IoU (70.4\%) is 21.4~pp higher. TriLICT nevertheless exceeds CLFT-Base at night by 3.0~pp mIoU, with gains on both pedestrians (3.9~pp) and cars (2.0~pp). In daytime, CLFT-Base reaches 79.0\% mIoU, 1.4~pp below TriLICT, again mainly through a lower People~IoU (64.8\% vs.\ 66.7\%). Both methods build on ViT encoders and DPT-style decoders and are trained on the same data, so the gap is consistent with the richer LiDAR input of TriLICT, native multi-channel LiDAR images registered to the camera view, compared with the projected point coordinates used by CLFT.

Fig.~\ref{fig:method_comparison} illustrates these results. In daytime, all five methods produce nearly identical masks, in line with their mIoU differences of at most 2.6~pp. At night, the three camera-only networks fragment or over-extend the pedestrian in the third column and miss the car on its left or label it as a pedestrian; in the fourth column they predict pedestrians in empty regions while missing the annotated one. Both camera--LiDAR methods segment the pedestrians and cars in both nighttime frames, and the masks of TriLICT follow the annotated extents more closely than those of CLFT-Base, most visibly for the pedestrian in the third column.

\subsection{LiDAR Channel Combination and Backbone Comparison}
\label{subsec:main_results}

Table~\ref{tab:main_results} reports all 16 dual-fusion models. The numerically best configuration is ViT-Large/16 with $\mathcal{C}_2$ in daytime (81.5\%) and ViT-Base/16 with $\mathcal{C}_1$ at night (74.8\%), but the spread is small: all models lie within 79.5--81.5\% mIoU in daytime and 73.2--74.8\% at night.

\begin{table*}[!t]
\centering
\captionsetup{font=footnotesize}
\caption{Segmentation performance (IoU, \%) of TriLICT under dual-fusion modality across all LiDAR channel combinations and backbone variants.
Bold: best per column; underlined: second best (ranked on unrounded values).}
\label{tab:main_results}
\resizebox{0.98\textwidth}{!}{
\begin{tabular}{llccc|ccc}
\toprule
\multirow{2}{*}{\textbf{Backbone}} & \multirow{2}{*}{\textbf{LiDAR Combo}} &
    \multicolumn{3}{c|}{\textbf{Daytime}} &
    \multicolumn{3}{c}{\textbf{Nighttime}} \\
\cmidrule(lr){3-5}\cmidrule(lr){6-8}
& & Cars IoU & People IoU & mIoU & Cars IoU & People IoU & mIoU \\
\midrule
\multirow{4}{*}{ViT-Base/16}
  & $\mathcal{C}_1$: range, nearir, signal   & \textbf{94.2} & 66.5 & 80.3 & 75.1 & \textbf{74.5} & \textbf{74.8} \\
  & $\mathcal{C}_2$: range, nearir, reflec   & 94.2 & 66.7 & 80.4 & 73.7 & \underline{74.3} & 74.0 \\
  & $\mathcal{C}_3$: range, signal, reflec   & 94.2 & 64.9 & 79.5 & \textbf{75.3} & 73.9 & \underline{74.6} \\
  & $\mathcal{C}_4$: nearir, signal, reflec  & \underline{94.2} & 67.1 & 80.7 & \underline{75.1} & 73.8 & 74.4 \\
\midrule
\multirow{4}{*}{ViT-Large/16}
  & $\mathcal{C}_1$: range, nearir, signal   & 94.1 & 68.6 & \underline{81.4} & 75.0 & 73.5 & 74.2 \\
  & $\mathcal{C}_2$: range, nearir, reflec   & 94.2 & \textbf{68.7} & \textbf{81.5} & 74.1 & 74.0 & 74.1 \\
  & $\mathcal{C}_3$: range, signal, reflec   & 94.0 & \underline{68.7} & 81.4 & 72.9 & 73.6 & 73.2 \\
  & $\mathcal{C}_4$: nearir, signal, reflec  & 94.2 & 67.7 & 81.0 & 74.4 & 72.6 & 73.5 \\
\bottomrule
\end{tabular}
}
\vspace{-1.5em}
\end{table*}

In daytime, Cars~IoU is virtually constant (94.0--94.2\%) and all variation comes from People~IoU (64.9--68.7\%), whereas at night the two classes reach comparable IoU (72.6--75.3\%). The higher nighttime People~IoU does not make the night the easier scenario: nighttime pedestrians walk much closer to the robot and are far larger in the image (median mask area of 3{,}528 vs.\ 93 pixels), while the registered LiDAR image covers on average only 75.2\% of the camera image at night, compared with 87.5\% in daytime.

Scaling from ViT-Base/16 to ViT-Large/16 helps only in daytime and only for pedestrians: averaged over the four combinations, daytime mIoU rises by 1.1~pp (People~IoU $+2.1$~pp, Cars~IoU unchanged). At night ViT-Large/16 brings no benefit (73.8\% vs.\ 74.5\%) and its runs early-stop sooner (best epochs 31--50 vs.\ 57--67), suggesting that nighttime performance is limited by the diversity of the single-session training data rather than by model capacity.

Because each $\mathcal{C}_k$ omits exactly one channel, Table~\ref{tab:main_results} can be read directly as a leave-one-out channel ablation: the mIoU of $\mathcal{C}_k$ measures performance without the channel it excludes, so the lowest mIoU in a row identifies the most valuable channel. Table~\ref{tab:loo} reorganizes the mIoU columns accordingly and compares the spread across combinations with the run-to-run standard deviation of the RGB-only baseline.

\begin{table}[!t]
\centering
\captionsetup{font=footnotesize}
\caption{Leave-one-out reading of Table~\ref{tab:main_results}: mIoU (\%) obtained when the named channel is the one omitted. Lowest value per row (bold) identifies the most valuable channel in that setting. Spread: best minus worst combination; $\sigma_\text{RGB}$: seed standard deviation of the RGB-only baseline (Table~\ref{tab:modality}).}
\label{tab:loo}
\resizebox{\linewidth}{!}{
\begin{tabular}{llcccccc}
\toprule
\multirow{2}{*}{\textbf{Backbone}} & \multirow{2}{*}{\textbf{Scenario}} & \multicolumn{4}{c}{\textbf{Omitted channel}} & \multirow{2}{*}{Spread} & \multirow{2}{*}{$\sigma_\text{RGB}$} \\
\cmidrule(lr){3-6}
 & & reflec ($\mathcal{C}_1$) & signal ($\mathcal{C}_2$) & nearir ($\mathcal{C}_3$) & range ($\mathcal{C}_4$) & & \\
\midrule
\multirow{2}{*}{ViT-Base}
  & Daytime   & 80.3 & 80.4 & \textbf{79.5} & 80.7 & 1.2 & 0.4 \\
  & Nighttime & 74.8 & \textbf{74.0} & 74.6 & 74.4 & 0.8 & 0.9 \\
\midrule
\multirow{2}{*}{ViT-Large}
  & Daytime   & 81.4 & 81.5 & 81.4 & \textbf{81.0} & 0.5 & 0.3 \\
  & Nighttime & 74.2 & 74.1 & \textbf{73.2} & 73.5 & 1.0 & 2.5 \\
\bottomrule
\end{tabular}
}
\end{table}

Read in this way, the four combinations are almost indistinguishable. Their spread is only 0.5--1.2~pp mIoU per setting, comparable to the seed standard deviation of the RGB-only baseline and, at night, smaller than it, and the most valuable channel changes with backbone and scenario (nearir in daytime and signal at night for ViT-Base/16; range and nearir for ViT-Large/16). Only ViT-Base/16 in daytime exceeds two standard deviations, owing to a lower People~IoU of $\mathcal{C}_3$ that does not recur under ViT-Large/16.

This insensitivity has a physical explanation. Range, signal, and reflectivity come from the active laser return and are independent of ambient light, and every combination contains at least two of them. The passive near-infrared channel, by contrast, is zero on 54.7\% of the valid LiDAR pixels at night, yet combinations with and without it perform on par. Omitting range ($\mathcal{C}_4$) is not consistently harmful either, plausibly because the signal--reflectivity pair implicitly encodes distance through the $1/r^{2}$ falloff that signal retains and reflectivity compensates (Section~\ref{subsec:lidar_rep}). We keep $\mathcal{C}_2$ for the comparisons, as it gives the best daytime result and stays within 0.8~pp of the best combination in every setting.

\subsection{Full Evaluation Metrics}
\label{subsec:detailed}

Table~\ref{tab:detailed} reports the complete per-class and all-class metrics of ViT-Base/16 with $\mathcal{C}_2$ on both test splits. Foreground metrics (IoU, Precision, Recall, F1, AP) are computed over the two foreground classes; all-class metrics (PA, mAcc, FWIoU) include background.

\begin{table}[!t]
\centering
\captionsetup{font=footnotesize}
\caption{Complete evaluation metrics (\%) of TriLICT (ViT-Base/16, $\mathcal{C}_2$ \{range, nearir, reflec\}, dual-fusion).}
\label{tab:detailed}
\resizebox{\linewidth}{!}{
\begin{tabular}{llccccc}
\toprule
\textbf{Split} & \textbf{Class} & IoU & Prec & Rec & F1 & AP \\
\midrule
\multirow{3}{*}{Daytime}
  & Cars & 94.2 & 96.5 & 97.5 & 97.0 & 99.6 \\
  & People & 66.7 & 82.0 & 78.1 & 80.0 & 86.7 \\
  & Mean & 80.4 & 89.3 & 87.8 & 88.5 & 93.2 \\
\midrule
\multirow{3}{*}{Nighttime}
  & Cars & 73.7 & 83.5 & 86.2 & 84.9 & 92.8 \\
  & People & 74.3 & 79.8 & 91.4 & 85.2 & 91.3 \\
  & Mean & 74.0 & 81.7 & 88.8 & 85.0 & 92.0 \\
\midrule
\multicolumn{2}{l}{Daytime: PA / mAcc / FWIoU}    & \multicolumn{5}{l}{99.8\% \;/\; 91.9\% \;/\; 99.6\%} \\
\multicolumn{2}{l}{Nighttime: PA / mAcc / FWIoU}  & \multicolumn{5}{l}{95.9\% \;/\; 91.7\% \;/\; 92.6\%} \\
\bottomrule
\end{tabular}
}
\vspace{-1.0em}
\end{table}

In daytime, car segmentation is close to saturation and well balanced (Precision 96.5\%, Recall 97.5\%, AP 99.6\%), while for pedestrians Precision (82.0\%) exceeds Recall (78.1\%), i.e.\ misses are slightly more frequent than false alarms, consistent with their small size. At night the pattern reverses: Recall exceeds Precision for both classes (Cars 86.2\% vs.\ 83.5\%, People 91.4\% vs.\ 79.8\%), so TriLICT rarely misses an annotated object but tends to over-extend its masks. The nighttime pedestrian AP of 91.3\% compares with only 65.9\% for the ViT-Base/16 RGB-only baseline. PA and FWIoU are dominated by the background class, which covers 97.5\% of the daytime and 86.8\% of the nighttime test pixels; mIoU over the foreground classes is therefore the more informative metric.

\subsection{Computational Cost}
\label{subsec:complexity}

\begin{table}[!t]
\centering
\captionsetup{font=footnotesize}
\caption{Model size, computational cost, and inference speed at $384\times384$ input resolution. FLOPs are multiply--accumulate operations of one forward pass; FPS is measured on one NVIDIA RTX~3090 with batch size 1 in FP32 (median of three runs).}
\label{tab:complexity}
\resizebox{0.98\linewidth}{!}{
\begin{tabular}{llccc}
\toprule
\textbf{Method} & \textbf{Input} & Params (M) & FLOPs (G) & FPS \\
\midrule
RGB-only (ViT-Base/16)  & C   & 103.7 & 104.6 & 63.8 \\
RGB-only (ViT-Large/16) & C   & 325.5 & 242.5 & 27.1 \\
SegFormer-B2            & C   & 27.3  & 33.1  & 98.6 \\
DeepLabV3+ (ResNet-101) & C   & 59.3  & 49.8  & 124.9 \\
\midrule
TriLICT (ViT-Base/16)   & C+L & 202.4 & 178.5 & 35.2 \\
TriLICT (ViT-Large/16)  & C+L & 646.0 & 454.2 & 13.5 \\
\bottomrule
\end{tabular}
}
\end{table}

As shown in Table~\ref{tab:complexity}, the LiDAR stream roughly doubles the parameters of the RGB-only model and raises its FLOPs by a factor of 1.7 (ViT-Base/16) and 1.9 (ViT-Large/16), mainly through the second encoder. TriLICT runs at 35.2~FPS with ViT-Base/16 and 13.5~FPS with ViT-Large/16, both above the 10~Hz LiDAR frame rate. As ViT-Base/16 matches ViT-Large/16 at night and trails it by only about 1~pp in daytime, while running 2.6$\times$ faster, it offers the better accuracy--efficiency trade-off. SegFormer-B2 and DeepLabV3+ are much lighter, but their efficiency does not translate into usable nighttime accuracy.

%% file: sec/06_Conclusion.tex
\section{Conclusion}\label{sec:conclusion}

In this paper, we presented TriLICT, a dual-stream transformer for semantic segmentation that fuses RGB imagery with native multi-channel LiDAR images registered to the camera view. Across four LiDAR channel combinations and two ViT backbones, evaluated on daytime and real nighttime data, TriLICT with ViT-Base/16 and range, near-infrared, and reflectivity reaches 74.0\% mIoU at night, against 52.6\% for the RGB-only baseline and 56.2\% for the strongest camera-only network, SegFormer-B2, with pedestrian IoU rising from 41.6\% to 74.3\%. In daytime it reaches 80.4\%, on par with its RGB-only baseline (80.6\%) and DeepLabV3+ (80.0\%) and 1.2~pp below SegFormer-B2 (81.6\%), so the LiDAR stream provides nighttime robustness without a daytime cost.

Our experiments further indicate that the choice of LiDAR channels is a second-order factor: all four combinations perform within about 1~pp mIoU, since each retains at least two active, illumination-invariant channels. TriLICT with ViT-Base/16 offers the best accuracy--efficiency trade-off (74.0\% nighttime mIoU at 35~FPS). These findings support native LiDAR images as a practical input for image-pretrained transformers.

In future work, we plan to elevate the TriLICT framework from semantic segmentation to panoptic segmentation, providing a unified understanding of both background context and foreground object instances. Furthermore, we intend to investigate the robustness of native multi-channel LiDAR fusion under adverse weather conditions, such as rain and snow, and explore temporal feature fusion to improve perception stability in highly dynamic driving scenarios.